\documentclass[letterpaper]{article} 
\usepackage{aaai24}  
\usepackage{times}  
\usepackage{helvet}  
\usepackage{courier}  
\usepackage[hyphens]{url}  
\usepackage{graphicx} 
\usepackage{natbib}  
\usepackage{caption} 
\usepackage{algorithm}
\usepackage{algorithmic}
\usepackage{amsthm,amsmath,amssymb}
\usepackage{bm}
\DeclareMathOperator*{\argmax}{argmax}

\usepackage{multirow}   
\usepackage{booktabs} 
\usepackage{makecell}
\usepackage{bbding}
\usepackage{nicematrix}
\usepackage{threeparttable}
\usepackage{amsfonts}
\usepackage{subcaption}
\usepackage{newfloat}
\usepackage{listings}
\DeclareCaptionStyle{ruled}{labelfont=normalfont,labelsep=colon,strut=off} 
\floatstyle{ruled}
\newfloat{listing}{tb}{lst}{}
\floatname{listing}{Listing}
\title{Unsupervised Cross-Domain Image Retrieval via Prototypical Optimal Transport}
\author{
    Bin Li\textsuperscript{\rm 1},
    Ye Shi\textsuperscript{\rm 1},
    Qian Yu\textsuperscript{\rm 2},
    Jingya Wang\textsuperscript{\rm 1}\thanks{Corresponding author.} 
}
\affiliations{
    \textsuperscript{\rm 1}ShanghaiTech University\\
    \textsuperscript{\rm 2}Beihang University\\
    
    \{libin3, shiye, wangjingya\}@shanghaitech.edu.cn; qianyu@buaa.edu.cn
}

\usepackage{bibentry}

\begin{document}

\maketitle

\begin{abstract}
Unsupervised cross-domain image retrieval (UCIR) aims to retrieve images sharing the same category across diverse domains without relying on labeled data. Prior approaches have typically decomposed the UCIR problem into two distinct tasks: intra-domain representation learning and cross-domain feature alignment. However, these segregated strategies overlook the potential synergies between these tasks. This paper introduces ProtoOT, a novel Optimal Transport formulation explicitly tailored for UCIR, which integrates intra-domain feature representation learning and cross-domain alignment into a unified framework. ProtoOT leverages the strengths of the K-means clustering method to effectively manage distribution imbalances inherent in UCIR. By utilizing K-means for generating initial prototypes and approximating class marginal distributions, we modify the constraints in Optimal Transport accordingly, significantly enhancing its performance in UCIR scenarios. Furthermore, we incorporate contrastive learning into the ProtoOT framework to further improve representation learning. This encourages local semantic consistency among features with similar semantics, while also explicitly enforcing separation between features and unmatched prototypes, thereby enhancing global discriminativeness. ProtoOT surpasses existing state-of-the-art methods by a notable margin across benchmark datasets. Notably, on DomainNet, ProtoOT achieves an average P@200 enhancement of 18.17\%, and on Office-Home, it demonstrates a P@15 improvement of 3.83\%. 

\end{abstract}

\section{Introduction}

Cross-domain image retrieval (CIR) aims at utilizing imagery data from one domain as queries to retrieve relevant samples from distinct domains. While significant progress has been made in supervised CIR studies \cite{huang2015cross,ji2017cross}, these advancements are constrained by the requirement for annotated labels. Unfortunately, this reliance on manual labels hampers the practical scalability of supervised CIR methods, due to the considerable costs associated with label acquisition in real-world scenarios. As a result, the usability of existing supervised CIR techniques becomes limited.

\begin{figure}[htpb]
\includegraphics[width=0.47\textwidth]{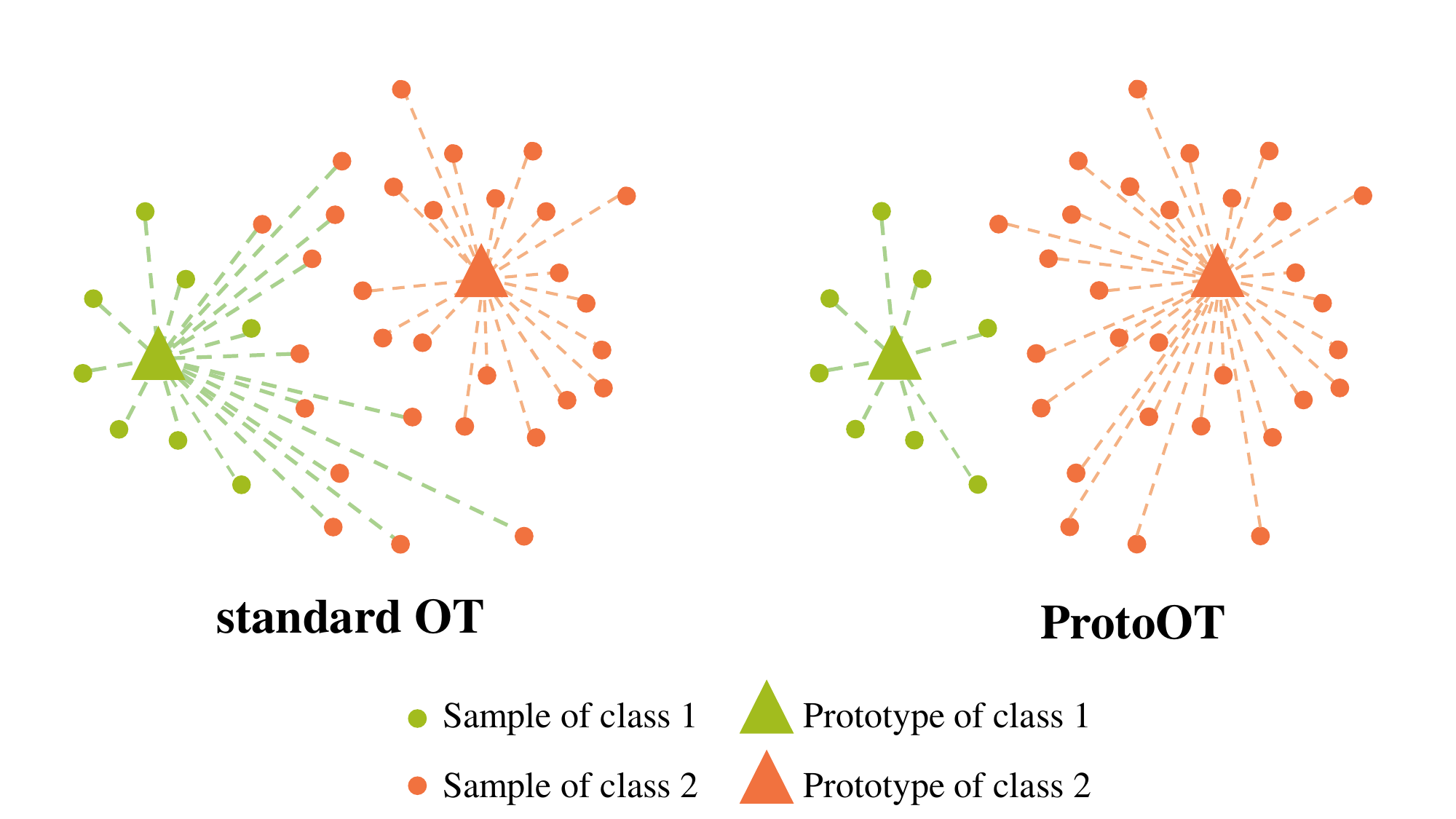}
\caption{
{Comparison between the standard OT and our proposed ProtoOT to deal with distribution imbalance. Different colors represent different categories, and the dashed lines indicate that the samples are matched to related prototypes.} 
}
\label{fig: the standard OT and ProtoOT.}
\end{figure}

To address this challenge, recent studies \cite{hu2022feature,wang2023correspondence} have explored unsupervised cross-domain image retrieval (UCIR) without depending on labeled information. However, the performance of these UCIR methods tends to be suboptimal due to the intricate nature of domain-specific clustering representation learning and the complexities involved in accommodating noise-tolerant cross-domain associations. Moreover, these studies tend to treat UCIR as two distinct tasks: intra-domain representation learning and cross-domain feature alignment. Regrettably, this segregated approach overlooks the underlying correlations and potential synergies between these tasks, resulting in subpar performance. 

In this paper, we investigate the UCIR problem by seamlessly integrating the intra-domain feature representation learning and cross-domain alignment. In particular, the former provides representative features for the alignment procedure in the latter, while the latter further enhances the representation in the former by integrating cross-domain knowledge. Notably, the core of both intra-domain representation learning and cross-domain alignment lies in minimizing the distance between the sample distribution and prototype distribution in order to obtain features that are both intra-class coherent and inter-class distinctive. Recent years have seen great success in applying Optimal Transport (OT) \cite{villani2009optimal} in both representation learning \cite{caron2020unsupervised,asano2019self} and cross-domain alignment \cite{flamary2016optimal,courty2017joint}. This naturally motivates us to a unified OT method for both intra-domain representation learning and cross-domain alignment for the UCIR problem. However, the UCIR task presents challenges stemming from significant imbalances in both intra-domain and cross-domain distributions. The standard OT methods lead to balanced assignments that are not suitable for UCIR, as shown in Figure \ref{fig: the standard OT and ProtoOT.}. Existing unbalanced OT methods, such as \cite{sejourne2022unbalanced,zhan2021unbalanced} also fall short in addressing UCIR's intricacies. A critical issue arises in existing unbalanced OT methods where some samples may remain unassigned to prototypes, consequently hindering the training process of UCIR.

To address this issue, we develop a novel OT formulation named Prototypical Optimal Transport (ProtoOT), tailored explicitly for UCIR. ProtoOT leverages the strengths of the K-means clustering method to effectively handle distribution imbalances in UCIR. By utilizing K-means for generating initial prototypes and approximating class marginal distributions, we modify the constraints in OT accordingly, significantly enhancing its performance in UCIR scenarios. To further enhance the representation learning, we employ contrastive learning \cite{oord2018representation} within the ProtoOT framework. This encourages local semantic consistency among features with similar semantics, while also explicitly enforcing separation between features and unmatched prototypes, thereby enhancing global discriminativeness. Our main contributions are summarized as follows: 
\begin{itemize}
    \item We address the UCIR problem by synergistically tackling intra-domain feature representation learning and cross-domain assignment, presenting the first unified Optimal Transport framework that simultaneously addresses both tasks.
    \item We introduce ProtoOT, a novel OT formulation that effectively balances the strengths of K-means and OT to handle substantial distribution imbalances in the context of UCIR. 
    \item ProtoOT outperforms existing state-of-the-art techniques by a substantial margin across benchmark datasets. Notably, on DomainNet, ProtoOT achieves an average enhancement of 18.17\% in terms of P@200, and on Office-Home, an improvement of 3.83\% in terms of P@15. 
\end{itemize}

\section{Related Work}

\subsection{Cross-Domain Image Retrieval}
Cross-domain image retrieval (CIR) stands as a well-explored task in computer vision. It involves retrieving images from one visual domain based on a query image from a different domain. CIR boasts extensive practical applications; in the fashion domain, for example\cite{gajic2018cross,bao2022mmfl}, it finds utility in matching user-provided images with a product database, thus presenting users with target products or accessories. Most existing works\cite{sain2021stylemeup,ji2017cross,lee2018cross} heavily lean on well-annotated data within the supervised learning framework, yet sacrifice scalability for real-world CIR implementations.
Recently, there have been efforts\cite{hu2022feature,wang2023correspondence} to move beyond relying on labels and tackle a tougher challenge known as unsupervised cross-domain image retrieval (UCIR). To bridge the gap between different domains, the preceding works\cite{hu2022feature,wang2023correspondence} split the UCIR task into intra-domain representation learning and cross-domain alignment, designing different algorithms for each part. However, they overlooked the essential correlations between these two parts, making it difficult for them to mutually enhance each other.

\subsection{Unsupervised Domain Adaptation}
Unsupervised domain adaptation (UDA) aims to address the challenge of transferring knowledge from a labeled source domain to a target domain without any annotations. To mitigate the domain shift, most existing methods have focused on reducing domain discrepancy through techniques like Maximum Mean Discrepancy \cite{gretton2012kernel,long2017deep}, domain adversarial training \cite{zhang2018collaborative,dai2020adversarial} and prototypes alignment\cite{lin2022prototype,li2022unsupervised,yue2021prototypical,chang2022unified}. However, while these methods have demonstrated substantial efficacy in UDA, they cannot be used directly for UCIR tasks because there are no labels in the source domain. In this paper, we delve into a more challenging scenario in which both the source and target domains are unlabeled. The problem of exploring weak self-representation learning for both domains while concurrently transferring knowledge from the other remains unresolved. 

\subsection{Optimal Transport}
Optimal transport (OT), initially proposed by Kantorovich \cite{villani2009optimal}, offers an efficient solution for transferring mass from one distribution to another. 
The development of optimized solvers\cite{cuturi2013sinkhorn, lahn2024combinatorial} have provided a guarantee for the wide application of OT. In the field of computer vision, OT has been utilized in diverse tasks like point cloud registration\cite{shen2021accurate} and learning with noisy labels\cite{chang2023csot}. These tasks involve seeking an optimal mapping between data distributions to minimize a cost function. Notably, \cite{asano2019self,caron2020unsupervised} have achieved significant advancements in unsupervised representation learning through OT-driven matching of samples to prototypes. Furthermore, in the context of domain adaptation, OT-based approaches\cite{courty2014domain,courty2017joint,redko2019optimal,chang2022unified} have exhibited promise in mitigating the gap between source and target domains by aligning their respective probability distributions. However, jointly exploring self-supervised feature representation learning and domain adaptation within a unified OT framework remains unexplored. 

\begin{figure*}[htpb]
\centering
\includegraphics[width=0.95\textwidth]{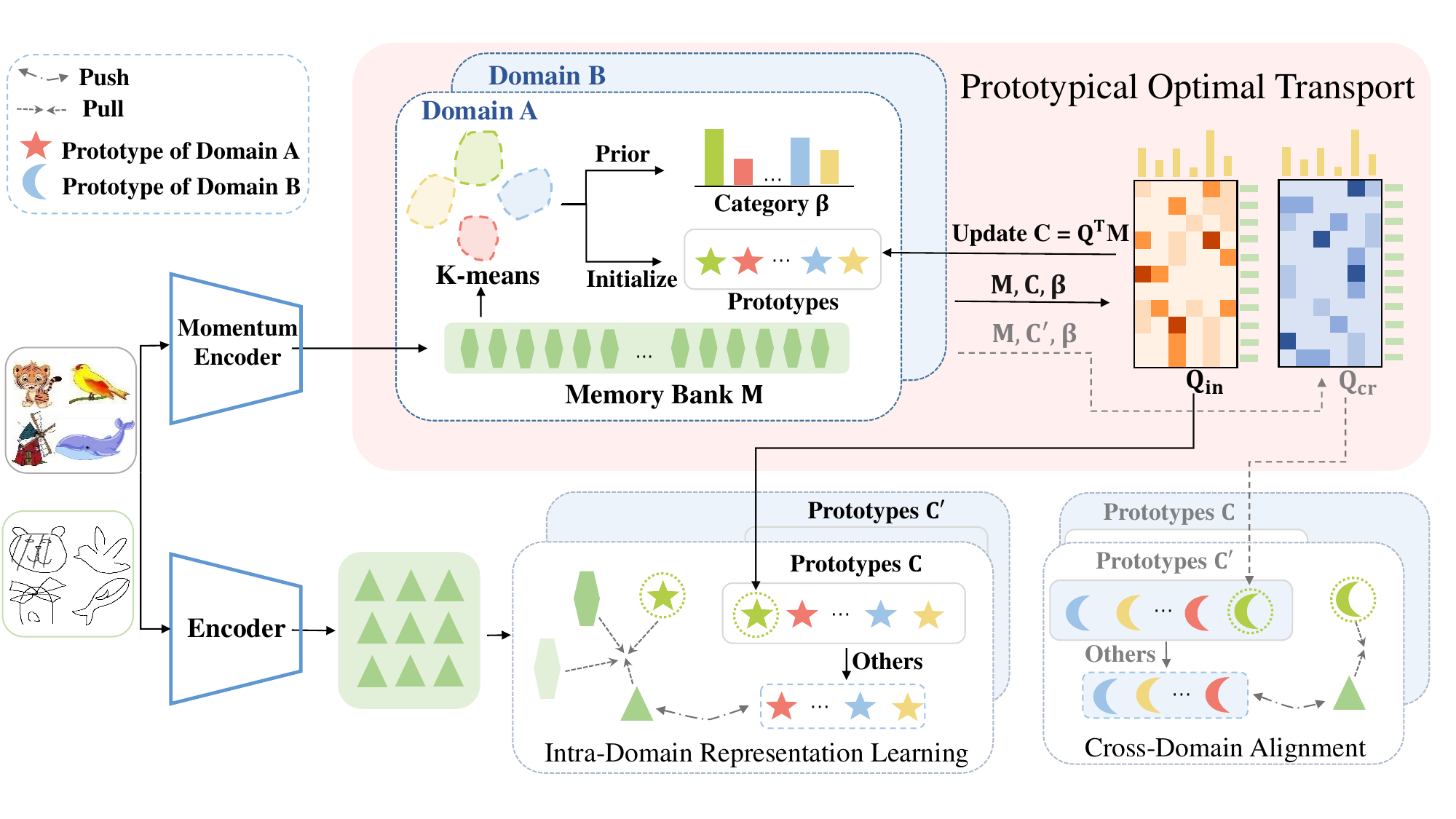}
\caption{
{An overview of the proposed method for UCIR. We utilize K-means to generate initial prototypes and approximate class marginal distributions, which have modified the constraints of Optimal Transport. Both the intra-domain representation learning and cross-domain alignment are based on ProtoOT. Furthmore, the two are committed to enhancing the local cosistency and global discriminability of features by employing the same form of contrastive loss. }
}
\label{fig:overview}
\end{figure*}

\section{Methodology}

In the UCIR task, we deal with two distinct domains, denoted as $\mathcal{D}_{A} = \{(\bm{x}_i^A)\}_{i=1}^{N_A}$ and $\mathcal{D}_{B} = \{(\bm{x}_j^B)\}_{j=1}^{N_B}$, both of which lack labeled data. The goal is to retrieve images of the category $k$ in domain $B$ based on a query image of category $k$ in domain $A$. To address this challenging task,
in this paper, we propose a unified OT framework, i.e. ProtoOT from two perspectives, i.e., intra-domain clustering and cross-domain alignment, as shown in Figure \ref{fig:overview}.

\subsection{ProtoOT}
Optimal Transport focuses on seeking the most efficient way to transform the mass of one distribution into another under the given marginal constraints. In unsupervised situations, where no information is available constraints in OT, SwAV \cite{caron2020unsupervised} adopts the standard OT to obtain the representations of mini-clusters. Given a similarity matrix $\bm{S}\in \mathbb{R}^{r\times c}$, the optimal coupling matrix $\bm{Q}^*$ of the standard OT can be obtained by solving the following optimization problem: 
\begin{equation}
        \bm{Q}^*(\bm{S}) =
            \argmax_{\bm{Q}\in\bm{U}}
                \text{Tr}(\bm{Q}^\top\bm{S})
                    +\varepsilon\textit{H}(\bm{Q}),
\end{equation}

\begin{equation}
    \bm{U}=\left\{\bm{Q}\in\mathbb{R}_{+}^{r \times c}
        |\bm{Q}\bm{1}_c=\frac{1}{r}\bm{1}_r,
        \bm{Q}^\top\bm{1}_r=\frac{1}{c}\bm{1}_c
        \right\}.
\end{equation}
where $\varepsilon >0$, the entropic regularization term $\textit{H}(\bm{Q})=-\sum_{ij}\bm{Q}_{ij}\log\bm{Q}_{ij}$.

Since the semantic information of the categories in UCIR is essential, enforcing uniform matching blindly is clearly unreasonable. To this end, we propose ProtoOT tailored for the UCIR task, which incorporates the results of K-means to the class marginal distribution for OT, instead of simply assuming a uniform distribution. 

\begin{equation}
        \bm{Q}^*(\bm{M},\bm{C},\bm{\beta}) =
            \argmax_{\bm{Q}\in\bm{U}(\bm{\beta})}
                \text{Tr}(\bm{Q}^\top\bm{M}\bm{C}^T)
                    +\varepsilon\textit{H}(\bm{Q}),
\end{equation}

\begin{equation}
    \bm{U}(\bm{\beta})=\left\{\bm{Q}\in\mathbb{R}_{+}^{r \times c}
        |\bm{Q}\bm{1}_c=\frac{1}{r}\bm{1}_r,
        \bm{Q}^\top\bm{1}_r=\bm{\beta}
        \right\}.
\end{equation}

\begin{equation}
    \bm{\beta} = [{{\beta}}_1, {{\beta}}_2,..., {{\beta}}_c ]^T
\end{equation}
\noindent where $\bm{M}$ is the set of the features, $\bm{C}$ is the prototypes, $\bm{1}_i$ is the $i$-dimensional vector of ones and ${\beta}_j\in \mathbb{R}$ is the probability of the $j$-th pseudo label generated by K-Means.

\subsection{Intra-Domain Representation Learning via ProtoOT}
Here we employ ProtoOT for intra-domain Representation Learning. First, we maintain a memory bank $\bm{M}$ by a momentum encoder $f_m$\cite{he2020momentum} for each domain. For each batch, we use the feature encoder $f$ to obtain features $\bm{q}$, whose augmented counterpart $\bm{k}$ encoded by $f_m$ can be found in the memory bank. Follow \cite{he2020momentum,chen2020improved}, we update the parameters using gradient backpropagation only within $f$, and then utilize the parameters of $f$ to momentum update $f_m$. We perform K-means on the memory bank $\bm{M}$ and use the obtained centroids from K-means as the initialized prototypes of intra-domain clustering. 

The UCIR task is a category-level cross-domain retrieval task, achieving a good intra-domain feature representation is important. The effectiveness of SwAV \cite{caron2020unsupervised} in learning representations of mini-clusters highlights the superiority of OT-based methods in clustering, which motivates us to utilize OT to assist in the learning of category semantic features. As previously mentioned, ProtoOT integrates the distribution information from K-means, allowing it to acquire a confident transport plan $\bm{Q}^*_{in}$ even in the absence of any data labels. Therefore, we use ProtoOT to acquire pseudo labels ${y}_i\in \bm{Y}$ and corresponding prototypes $\bm{C}_{y_i}$  for each sample, where $\bm{Y}$ is the set of all pseudo labels.
\begin{equation}
    {y}_i = \argmax_{{y}\in \bm{Y}} \bm{Q}^*_{in}(\bm{M},\bm{C},\bm{\beta}).
\end{equation}

To attain semantic clustering of features, aggregating features with local similarity is reasonable. However, merely focusing on the local similarity among features cannot effectively eliminate interference from features that are semantically similar yet not from the same category. This drives us to explore global discrimination as well. 
Specifically, to achieve local consistency in features, we bring together the feature $\bm{q}_i$ with its augmentation $\bm{k}_i$ in the embedding space. In addition to this, getting closer to the nearest neighbor $\bm{n}_i$ of the feature in the memory bank and converging with the matched prototype $\bm{C}_{y_i}$ is intuitive. We denote a set $\mathcal{P}_i^\text{in}$ to collect intra-domain positive triplets of sample $i$, i.e., $\mathcal{P}_i^\text{in}:=\{\bm{k}_i, \bm{n}_i, \bm{C}_{y_i}\}$.  
To explore global discrimination, it is crucial to ensure a large margin between prototypes of different categories. For this purpose, we consider prototypes that are not matched with features by ProtoOT as negative samples. Similarly, we define the set $\mathcal{N}^\text{in}_i =\{\bm{C}_y\in \bm{C} | {y}\in \bm{Y}, {y}\neq {y}_i\}$. 

Unlike K-means, which uses per-sample to obtain hard pseudo labels and form prototypes, here we utilize the transport plan $\bm{Q}^*_{in}(\bm{M},\bm{C},\bm{\beta})$ of ProtoOT  to get soft pseudo labels. This enables the prototypes $\bm{C}$ to be more globally representative, i.e.,
\begin{equation}
    \bm{C} = {\bm{Q}^*_{in}}^T\bm{M}.
\end{equation}
Then, we integrate local consistency and global discrimination into a contrastive learning loss:
\begin{equation}
    \mathcal{L}_{in} =\sum\limits_{i \in \mathcal{D}}
    -\log\frac{\exp(\bm{q}^T_i\bm{p}^\text{in}_i/\tau)}{\exp(\bm{q}^T_i\bm{p}^\text{in}_i/\tau)+\sum\limits_{\bm{n}\in \mathcal{N}^\text{in}_i} \exp(\bm{q}^T_i\bm{n}/\tau)},
\end{equation}
\noindent where $\bm{p}^\text{in}_i \in \mathcal{P}_i^\text{in}$ represents a positive sample of sample $i$, $\tau$ is the temperature value.

\begin{table*}[htpb]
\centering
\resizebox{\linewidth}{!}{
\begin{tabular}{cccccccccccccc}
\toprule
\hline
\multicolumn{2}{c}{\multirow{2}{*}{Method}}   & \multicolumn{3}{c|}{Clipart{$\rightarrow$}Sketch} & \multicolumn{3}{c|}{Sketch{$\rightarrow$}Clipart}  & \multicolumn{3}{c|}{Infograph{$\rightarrow$}Real} & \multicolumn{3}{c}{Real{$\rightarrow$}Infograph}   \\
\multicolumn{2}{c}{}                                          & \multicolumn{1}{c}{P@50} & \multicolumn{1}{c}{P@100}  & \multicolumn{1}{c|}{P@200} & \multicolumn{1}{c}{P@50} & \multicolumn{1}{c}{P@100}  & \multicolumn{1}{c|}{P@200} & \multicolumn{1}{c}{P@50} & \multicolumn{1}{c}{P@100}  & \multicolumn{1}{c|}{P@200} & \multicolumn{1}{c}{P@50} & \multicolumn{1}{c}{P@100}  & \multicolumn{1}{c}{P@200}
\\ 
\midrule

\multicolumn{2}{l}{ID~\cite{wu2018unsupervised}}          & \multicolumn{1}{c}{49.46}  & \multicolumn{1}{c}{46.09} & \multicolumn{1}{c|}{40.44}     & \multicolumn{1}{c}{54.38}  & \multicolumn{1}{c}{47.12} & \multicolumn{1}{c|}{37.73} & \multicolumn{1}{c}{28.27}  & \multicolumn{1}{c}{27.44} & \multicolumn{1}{c|}{26.33} & \multicolumn{1}{c}{39.98}  & \multicolumn{1}{c}{31.77} & \multicolumn{1}{c}{24.84}  \\ 

\multicolumn{2}{l}{\multirow{1}{*}{ProtoNCE~\cite{li2020prototypical}}}        & \multicolumn{1}{c}{46.85}  & \multicolumn{1}{c}{42.67} & \multicolumn{1}{c|}{36.35}     & \multicolumn{1}{c}{54.52}  & \multicolumn{1}{c}{45.04} & \multicolumn{1}{c|}{35.06} & \multicolumn{1}{c}{28.41}  & \multicolumn{1}{c}{28.53} & \multicolumn{1}{c|}{28.50} & \multicolumn{1}{c}{57.01}  & \multicolumn{1}{c}{41.84} & \multicolumn{1}{c}{30.33}  \\

\multicolumn{2}{l}{\multirow{1}{*}{CDS~\cite{kim2021cds}}}        & \multicolumn{1}{c}{45.84}  & \multicolumn{1}{c}{42.37} & \multicolumn{1}{c|}{37.16}     & \multicolumn{1}{c}{59.13}  & \multicolumn{1}{c}{48.83} & \multicolumn{1}{c|}{37.40} & \multicolumn{1}{c}{28.51}  & \multicolumn{1}{c}{27.92} & \multicolumn{1}{c|}{27.48} & \multicolumn{1}{c}{56.69}  & \multicolumn{1}{c}{39.76} & \multicolumn{1}{c}{26.38}  \\

\multicolumn{2}{l}{\multirow{1}{*}{PCS~\cite{yue2021prototypical}}}               & \multicolumn{1}{c}{51.01}  & \multicolumn{1}{c}{46.87} & \multicolumn{1}{c|}{40.19}     & \multicolumn{1}{c}{59.70}  & \multicolumn{1}{c}{50.67} & \multicolumn{1}{c|}{39.38} & \multicolumn{1}{c}{30.56}  & \multicolumn{1}{c}{30.27} & \multicolumn{1}{c|}{29.68} & \multicolumn{1}{c}{55.42}  & \multicolumn{1}{c}{42.13} & \multicolumn{1}{c}{30.76}  \\

\multicolumn{2}{l}{\multirow{1}{*}{DD~\cite{hu2022feature}}}              & \multicolumn{1}{c}{56.31}  & \multicolumn{1}{c}{52.74} & \multicolumn{1}{c|}{47.38}     & \multicolumn{1}{c}{63.07}  & \multicolumn{1}{c}{57.26} & \multicolumn{1}{c|}{48.17} & \multicolumn{1}{c}{35.52}  & \multicolumn{1}{c}{35.24} & \multicolumn{1}{c|}{34.35} & \multicolumn{1}{c}{57.74}  & \multicolumn{1}{c}{46.69} & \multicolumn{1}{c}{35.47}  \\

\multicolumn{2}{l}{\multirow{1}{*}{\textbf{ProtoOT}}}              
& \multicolumn{1}{c}{\textbf{70.46}}& \multicolumn{1}{c}{\textbf{69.41}}     & \multicolumn{1}{c|}{\textbf{67.41}}    & \multicolumn{1}{c}{\textbf{82.79}}    & \multicolumn{1}{c}{\textbf{78.68}}    & \multicolumn{1}{c|}{\textbf{71.31}}    & \multicolumn{1}{c}{\textbf{40.65}}    & \multicolumn{1}{c}{\textbf{40.35}}    & \multicolumn{1}{c|}{\textbf{40.05}}    & \multicolumn{1}{c}{\textbf{77.02}}    & \multicolumn{1}{c}{\textbf{67.33}}    &  \multicolumn{1}{c}{\textbf{49.41}}     \\

\midrule

\multicolumn{2}{c}{\multirow{2}{*}}  & \multicolumn{3}{c|}{Infograph{$\rightarrow$}Sketch} & \multicolumn{3}{c|}{Sketch{$\rightarrow$}Infograph}  & \multicolumn{3}{c|}{Painting{$\rightarrow$}Clipart} & \multicolumn{3}{c}{Clipart{$\rightarrow$}Painting}   \\
\multicolumn{2}{c}{}        & \multicolumn{1}{c}{P@50} & \multicolumn{1}{c}{P@100}  & \multicolumn{1}{c|}{P@200} & \multicolumn{1}{c}{P@50} & \multicolumn{1}{c}{P@100}  & \multicolumn{1}{c|}{P@200} & \multicolumn{1}{c}{P@50} & \multicolumn{1}{c}{P@100}  & \multicolumn{1}{c|}{P@200} & \multicolumn{1}{c}{P@50} & \multicolumn{1}{c}{P@100}  & \multicolumn{1}{c}{P@200}
\\ 
\midrule

\multicolumn{2}{l}{ID~\cite{wu2018unsupervised}}                        & \multicolumn{1}{c}{30.35}  & \multicolumn{1}{c}{29.04} & \multicolumn{1}{c|}{26.55}     & \multicolumn{1}{c}{42.20}  & \multicolumn{1}{c}{34.94} & \multicolumn{1}{c|}{27.52} & \multicolumn{1}{c}{64.67}  & \multicolumn{1}{c}{54.41} & \multicolumn{1}{c|}{40.07} & \multicolumn{1}{c}{42.37}  & \multicolumn{1}{c}{39.61} & \multicolumn{1}{c}{35.56}   \\ 

\multicolumn{2}{l}{\multirow{1}{*}{ProtoNCE~\cite{li2020prototypical}}}         & \multicolumn{1}{c}{28.24}  & \multicolumn{1}{c}{26.79} & \multicolumn{1}{c|}{24.23}     & \multicolumn{1}{c}{39.83}  & \multicolumn{1}{c}{31.99} & \multicolumn{1}{c|}{24.77} & \multicolumn{1}{c}{55.44}  & \multicolumn{1}{c}{43.74} & \multicolumn{1}{c|}{32.59} & \multicolumn{1}{c}{39.13}  & \multicolumn{1}{c}{35.87} & \multicolumn{1}{c}{32.07}  \\

\multicolumn{2}{l}{\multirow{1}{*}{CDS~\cite{kim2021cds}}}          & \multicolumn{1}{c}{30.55}  & \multicolumn{1}{c}{29.51} & \multicolumn{1}{c|}{27.00}     & \multicolumn{1}{c}{46.27}  & \multicolumn{1}{c}{36.11} & \multicolumn{1}{c|}{27.33} & \multicolumn{1}{c}{63.15}  & \multicolumn{1}{c}{47.30} & \multicolumn{1}{c|}{32.93} & \multicolumn{1}{c}{37.75}  & \multicolumn{1}{c}{35.18} & \multicolumn{1}{c}{32.76}  \\

\multicolumn{2}{l}{\multirow{1}{*}{PCS~\cite{yue2021prototypical}}}           & \multicolumn{1}{c}{30.27}  & \multicolumn{1}{c}{28.36} & \multicolumn{1}{c|}{25.35}     & \multicolumn{1}{c}{42.58}  & \multicolumn{1}{c}{34.09} & \multicolumn{1}{c|}{25.91} & \multicolumn{1}{c}{63.47}  & \multicolumn{1}{c}{53.21} & \multicolumn{1}{c|}{41.68} & \multicolumn{1}{c}{48.83}  & \multicolumn{1}{c}{46.21} & \multicolumn{1}{c}{42.10}  \\

\multicolumn{2}{l}{\multirow{1}{*}{DD~\cite{hu2022feature}}}           & \multicolumn{1}{c}{31.29}  & \multicolumn{1}{c}{29.33} & \multicolumn{1}{c|}{26.54}     & \multicolumn{1}{c}{43.66}  & \multicolumn{1}{c}{36.14} & \multicolumn{1}{c|}{28.12} & \multicolumn{1}{c}{66.42}  & \multicolumn{1}{c}{56.84} & \multicolumn{1}{c|}{46.72} & \multicolumn{1}{c}{52.58}  & \multicolumn{1}{c}{50.10} & \multicolumn{1}{c}{46.11}    \\

\multicolumn{2}{l}{\multirow{1}{*}{\textbf{ProtoOT}}}              
& \multicolumn{1}{c}{\textbf{37.16}}& \multicolumn{1}{c}{\textbf{36.30}}     & \multicolumn{1}{c|}{\textbf{34.42}}    & \multicolumn{1}{c}{\textbf{63.59}}    & \multicolumn{1}{c}{\textbf{53.30}}    & \multicolumn{1}{c|}{\textbf{38.75}}    & \multicolumn{1}{c}{\textbf{90.21}}    & \multicolumn{1}{c}{\textbf{87.38}}    & \multicolumn{1}{c|}{\textbf{77.14}}    & \multicolumn{1}{c}{\textbf{71.13}}    & \multicolumn{1}{c}{\textbf{71.15}}    &  \multicolumn{1}{c}{\textbf{70.50}}     \\

\midrule

\multicolumn{2}{c}{\multirow{2}{*}}  & \multicolumn{3}{c|}{Painting{$\rightarrow$}Quickdraw} & \multicolumn{3}{c|}{Quickdraw{$\rightarrow$}Painting}  & \multicolumn{3}{c|}{Quickdraw{$\rightarrow$}Real} & \multicolumn{3}{c}{Real{$\rightarrow$}Quickdraw}   \\
\multicolumn{2}{c}{}                                          & \multicolumn{1}{c}{P@50} & \multicolumn{1}{c}{P@100}  & \multicolumn{1}{c|}{P@200} & \multicolumn{1}{c}{P@50} & \multicolumn{1}{c}{P@100}  & \multicolumn{1}{c|}{P@200} & \multicolumn{1}{c}{P@50} & \multicolumn{1}{c}{P@100}  & \multicolumn{1}{c|}{P@200} & \multicolumn{1}{c}{P@50} & \multicolumn{1}{c}{P@100}  & \multicolumn{1}{c}{P@200}
\\ 
\midrule

\multicolumn{2}{l}{ID~\cite{wu2018unsupervised}}              & \multicolumn{1}{c}{20.34}  & \multicolumn{1}{c}{19.59} & \multicolumn{1}{c|}{18.79}     & \multicolumn{1}{c}{21.12}  & \multicolumn{1}{c}{19.81} & \multicolumn{1}{c|}{18.48} & \multicolumn{1}{c}{28.27}  & \multicolumn{1}{c}{27.46} & \multicolumn{1}{c|}{26.32} & \multicolumn{1}{c}{23.45}  & \multicolumn{1}{c}{22.79} & \multicolumn{1}{c}{22.01}   \\ 

\multicolumn{2}{l}{\multirow{1}{*}{ProtoNCE~\cite{li2020prototypical}}}          & \multicolumn{1}{c}{21.63}  & \multicolumn{1}{c}{21.24} & \multicolumn{1}{c|}{20.56}     & \multicolumn{1}{c}{23.95}  & \multicolumn{1}{c}{22.84} & \multicolumn{1}{c|}{21.56} & \multicolumn{1}{c}{26.38}  & \multicolumn{1}{c}{25.70} & \multicolumn{1}{c|}{24.45} & \multicolumn{1}{c}{25.10}  & \multicolumn{1}{c}{24.81} & \multicolumn{1}{c}{23.78}  \\

\multicolumn{2}{l}{\multirow{1}{*}{CDS~\cite{kim2021cds}}}        & \multicolumn{1}{c}{18.75}  & \multicolumn{1}{c}{18.89} & \multicolumn{1}{c|}{17.88}     & \multicolumn{1}{c}{21.37}  & \multicolumn{1}{c}{21.44} & \multicolumn{1}{c|}{19.46} & \multicolumn{1}{c}{19.28}  & \multicolumn{1}{c}{19.14} & \multicolumn{1}{c|}{18.67} & \multicolumn{1}{c}{15.36}  & \multicolumn{1}{c}{15.57} & \multicolumn{1}{c}{15.82}  \\

\multicolumn{2}{l}{\multirow{1}{*}{PCS~\cite{yue2021prototypical}}}         & \multicolumn{1}{c}{25.12}  & \multicolumn{1}{c}{24.65} & \multicolumn{1}{c|}{23.80}     & \multicolumn{1}{c}{24.03}  & \multicolumn{1}{c}{23.24} & \multicolumn{1}{c|}{22.13} & \multicolumn{1}{c}{34.82}  & \multicolumn{1}{c}{33.92} & \multicolumn{1}{c|}{31.73} & \multicolumn{1}{c}{28.98}  & \multicolumn{1}{c}{28.85} & \multicolumn{1}{c}{28.16}  \\

\multicolumn{2}{l}{\multirow{1}{*}{\text{DD}~\cite{hu2022feature}}}         & \multicolumn{1}{c}{39.72}  & \multicolumn{1}{c}{38.59} & \multicolumn{1}{c|}{37.63}     & \multicolumn{1}{c}{33.45}  & \multicolumn{1}{c}{33.81} & \multicolumn{1}{c|}{34.29} & \multicolumn{1}{c}{42.79}  & \multicolumn{1}{c}{42.75} & \multicolumn{1}{c|}{42.70} & \multicolumn{1}{c}{41.90}  & \multicolumn{1}{c}{42.10} & \multicolumn{1}{c}{41.59}    \\ 

\multicolumn{2}{l}{\multirow{1}{*}{\textbf{ProtoOT}}}              
 & \multicolumn{1}{c}{\textbf{63.96}}& \multicolumn{1}{c}{\textbf{62.75}}     & \multicolumn{1}{c|}{\textbf{61.48}}    & \multicolumn{1}{c}{\textbf{60.01}}    & \multicolumn{1}{c}{\textbf{60.33}}    & \multicolumn{1}{c|}{\textbf{59.49}}    & \multicolumn{1}{c}{\textbf{60.32}}    & \multicolumn{1}{c}{\textbf{60.64}}    & \multicolumn{1}{c|}{\textbf{60.75}}    & \multicolumn{1}{c}{\textbf{55.17}}    & \multicolumn{1}{c}{\textbf{55.56}}    &  \multicolumn{1}{c}{\textbf{56.37}}    \\

\hline
\bottomrule
\end{tabular}
}

\caption{Unsupervised Cross-domain Retrieval Accuracy (\%) on DomainNet}
\label{tab:acc on domainnet}
\end{table*}

\subsection{Cross-Domain Alignment via ProtoOT}
Through intra-domain representation learning, prototypes with strong category semantics can be formed separately in domain $\mathcal{D}_{A}$ and domain $\mathcal{D}_{B}$. In order to eliminate the domain gap, it is crucial to facilitate the mutual transfer of knowledge learned in the intra-domain clustering between the two domains. The achievements of OT \cite{courty2014domain,courty2017joint} in domain adaptation lead us to believe that OT-based methods are capable of addressing this fully unsupervised adaptation between the two unlabeled domains. However, it requires simultaneous consideration of the potential data imbalanced in both domains, which makes the OT with enforced uniform matching even less suitable for the cases where imbalance is more severe. The natural tolerance of ProtoOT to data imbalance equips it with the ability to address this challenging problem.

Specifically, we utilize ProtoOT to match the features from domain $\mathcal{D}_{A}$ to the prototypes $\bm{C}'$of domain $\mathcal{D}_{B}$, where $\bm{Q}^*_{cr}$ is the cross-domain transport plan between the memory bank $\bm{M}$ of domain  $\mathcal{D}_{A}$ and prototypes $\bm{C}'$. 
\begin{equation}
    \bm{y}'_i = \argmax_{{y}'\in\bm{Y}'} \bm{Q}^*_{cr}(\bm{M}, \bm{C}', \bm{\beta}_{cr}), 
\end{equation}
\noindent where the symbol $'$ denotes a different domain,  $\bm{\beta}_{cr}$ is the categories distribution of $\bm{M}$ obtained by K-means. We consider the matched prototype $\bm{C}_{y'_i}$ as the positive sample and other prototypes are treated as negative samples $\bm n'\in \mathcal{N}^{cr}_i$. 
\begin{equation}
    \mathcal{N}^\text{cr}_i = \{\bm{C}'_{y'}\in \bm{C}'|{y'}\in \bm{Y}', {y'}\neq {y}'_i\}.
\end{equation}

To mitigate domain discrepancies, it is important to aggregate features that exhibit semantic similarity across domains while disregarding the impact of  other features that might share semantic resemblance but belong to different categories.
In light of this, to achieve cross-domain semantic alignment, we bring the matched prototypes from one domain closer to learn semantic consistency, while pushing away the unmatched prototypes from the other domain to attain discrimination across domains. Due to the large domain gap, relying on the nearest neighbor feature across domains based on similarity becomes unreasonable. The loss function $\mathcal{L}_{cr}$ of cross-domain alignment is as follows: 
\begin{equation}
    \mathcal{L}_{cr} =\sum\limits_{i \in \mathcal{D}}
    -\log\frac{\exp(\bm{q}_i^T\bm{C}'_{y'_i}/\tau)}{\exp(\bm{q}_i^T\bm{C}'_{y'_i}/\tau) + \sum\limits_{{n}'\in \mathcal{N}^\text{cr}_i} \exp(\bm{q}_i^T \bm{n}'/\tau)}.
\end{equation}

\begin{table*}[h]
\centering
\resizebox{\linewidth}{!}{
\begin{tabular}{cccccccccccccc}
\toprule
\hline
\multicolumn{2}{c}{\multirow{2}{*}{Method}}   & \multicolumn{3}{c|}{Art{$\rightarrow$}Real} & \multicolumn{3}{c|}{Real{$\rightarrow$}Art}  & \multicolumn{3}{c|}{Art{$\rightarrow$}Product} & \multicolumn{3}{c}{Product{$\rightarrow$}Art}   \\
\multicolumn{2}{c}{}                                          & \multicolumn{1}{c}{P@1} & \multicolumn{1}{c}{P@5}  & \multicolumn{1}{c|}{P@15} & \multicolumn{1}{c}{P@1} & \multicolumn{1}{c}{P@5}  & \multicolumn{1}{c|}{P@15} & \multicolumn{1}{c}{P@1} & \multicolumn{1}{c}{P@5}  & \multicolumn{1}{c|}{P@15} & \multicolumn{1}{c}{P@1} & \multicolumn{1}{c}{P@5}  & \multicolumn{1}{c}{P@15}
\\ 
\midrule

\multicolumn{2}{l}{ID~\cite{wu2018unsupervised}}               & \multicolumn{1}{c}{35.89}  & \multicolumn{1}{c}{33.13} & \multicolumn{1}{c|}{29.60}     & \multicolumn{1}{c}{39.89}  & \multicolumn{1}{c}{34.42} & \multicolumn{1}{c|}{27.65} & \multicolumn{1}{c}{25.88}  & \multicolumn{1}{c}{24.91} & \multicolumn{1}{c|}{22.49} & \multicolumn{1}{c}{32.17}  & \multicolumn{1}{c}{25.94} & \multicolumn{1}{c}{20.23}  \\ 

\multicolumn{2}{l}{\multirow{1}{*}{ProtoNCE~\cite{li2020prototypical}}}              & \multicolumn{1}{c}{40.50}  & \multicolumn{1}{c}{36.39} & \multicolumn{1}{c|}{34.00}     & \multicolumn{1}{c}{44.53}  & \multicolumn{1}{c}{39.26} & \multicolumn{1}{c|}{32.99} & \multicolumn{1}{c}{29.54}  & \multicolumn{1}{c}{27.89} & \multicolumn{1}{c|}{25.75} & \multicolumn{1}{c}{35.73}  & \multicolumn{1}{c}{30.61} & \multicolumn{1}{c}{24.55}  \\

\multicolumn{2}{l}{\multirow{1}{*}{CDS~\cite{kim2021cds}}}               & \multicolumn{1}{c}{45.08}  & \multicolumn{1}{c}{41.15} & \multicolumn{1}{c|}{38.73}     & \multicolumn{1}{c}{44.71}  & \multicolumn{1}{c}{40.75} & \multicolumn{1}{c|}{35.53} & \multicolumn{1}{c}{32.76}  & \multicolumn{1}{c}{31.47} & \multicolumn{1}{c|}{28.90} & \multicolumn{1}{c}{35.75}  & \multicolumn{1}{c}{32.48} & \multicolumn{1}{c}{26.82}  \\

\multicolumn{2}{l}{\multirow{1}{*}{PCS~\cite{yue2021prototypical}}}              & \multicolumn{1}{c}{41.70}  & \multicolumn{1}{c}{38.51} & \multicolumn{1}{c|}{36.22}     & \multicolumn{1}{c}{44.96}  & \multicolumn{1}{c}{39.88} & \multicolumn{1}{c|}{33.99} & \multicolumn{1}{c}{33.29}  & \multicolumn{1}{c}{31.50} & \multicolumn{1}{c|}{29.53} & \multicolumn{1}{c}{39.24}  & \multicolumn{1}{c}{34.77} & \multicolumn{1}{c}{28.77}  \\

\multicolumn{2}{l}{\multirow{1}{*}{DD~\cite{hu2022feature}}}            & \multicolumn{1}{c}{45.12}  & \multicolumn{1}{c}{42.33} & \multicolumn{1}{c|}{40.06}     & \multicolumn{1}{c}{47.95}  & \multicolumn{1}{c}{43.68} & \multicolumn{1}{c|}{38.38} & \multicolumn{1}{c}{35.39}  & \multicolumn{1}{c}{34.67} & \multicolumn{1}{c|}{32.61} & \multicolumn{1}{c}{42.51}  & \multicolumn{1}{c}{37.94} & \multicolumn{1}{c}{31.41}  \\

\multicolumn{2}{l}{\multirow{1}{*}{\textbf{ProtoOT}}}              
& \multicolumn{1}{c}{\textbf{47.38}}& \multicolumn{1}{c}{\textbf{45.49}}     & \multicolumn{1}{c|}{\textbf{43.52}}    & \multicolumn{1}{c}{\textbf{50.61}}    & \multicolumn{1}{c}{\textbf{46.64}}    & \multicolumn{1}{c|}{\textbf{41.51}}    & \multicolumn{1}{c}{\textbf{38.11}}    & \multicolumn{1}{c}{\textbf{36.50}}    & \multicolumn{1}{c|}{\textbf{35.10}}    & \multicolumn{1}{c}{\textbf{46.47}}    & \multicolumn{1}{c}{\textbf{41.63}}    &  \multicolumn{1}{c}{\textbf{34.47}}     \\ 

\midrule

\multicolumn{2}{c}{\multirow{2}{*}}  & \multicolumn{3}{c|}{Clipart{$\rightarrow$}Real} & \multicolumn{3}{c|}{Real{$\rightarrow$}Clipart}  & \multicolumn{3}{c|}{Product{$\rightarrow$}Real} & \multicolumn{3}{c}{Real{$\rightarrow$}Product}   \\
\multicolumn{2}{c}{}                                          & \multicolumn{1}{c}{P@1} & \multicolumn{1}{c}{P@5}  & \multicolumn{1}{c|}{P@15} & \multicolumn{1}{c}{P@1} & \multicolumn{1}{c}{P@5}  & \multicolumn{1}{c|}{P@15} & \multicolumn{1}{c}{P@1} & \multicolumn{1}{c}{P@5}  & \multicolumn{1}{c|}{P@15} & \multicolumn{1}{c}{P@1} & \multicolumn{1}{c}{P@5}  & \multicolumn{1}{c}{P@15}
\\ 
\midrule

\multicolumn{2}{l}{ID~\cite{wu2018unsupervised}}                      & \multicolumn{1}{c}{29.48}  & \multicolumn{1}{c}{26.48} & \multicolumn{1}{c|}{23.25}     & \multicolumn{1}{c}{35.51}  & \multicolumn{1}{c}{32.17} & \multicolumn{1}{c|}{27.96} & \multicolumn{1}{c}{50.73}  & \multicolumn{1}{c}{45.03} & \multicolumn{1}{c|}{39.05} & \multicolumn{1}{c}{45.12}  & \multicolumn{1}{c}{41.46} & \multicolumn{1}{c}{38.01}   \\ 

\multicolumn{2}{l}{\multirow{1}{*}{ProtoNCE~\cite{li2020prototypical}}}             & \multicolumn{1}{c}{25.25}  & \multicolumn{1}{c}{22.66} & \multicolumn{1}{c|}{20.83}     & \multicolumn{1}{c}{41.15}  & \multicolumn{1}{c}{37.66} & \multicolumn{1}{c|}{31.95} & \multicolumn{1}{c}{53.84}  & \multicolumn{1}{c}{48.25} & \multicolumn{1}{c|}{42.21} & \multicolumn{1}{c}{47.74}  & \multicolumn{1}{c}{44.85} & \multicolumn{1}{c}{41.21}  \\

\multicolumn{2}{l}{\multirow{1}{*}{CDS~\cite{kim2021cds}}}              & \multicolumn{1}{c}{32.51}  & \multicolumn{1}{c}{30.30} & \multicolumn{1}{c|}{27.80}     & \multicolumn{1}{c}{38.88}  & \multicolumn{1}{c}{36.48} & \multicolumn{1}{c|}{33.16} & \multicolumn{1}{c}{54.00}  & \multicolumn{1}{c}{50.07} & \multicolumn{1}{c|}{45.60} & \multicolumn{1}{c}{49.39}  & \multicolumn{1}{c}{47.27} & \multicolumn{1}{c}{43.98}  \\

\multicolumn{2}{l}{\multirow{1}{*}{PCS~\cite{yue2021prototypical}}}               & \multicolumn{1}{c}{29.07}  & \multicolumn{1}{c}{26.06} & \multicolumn{1}{c|}{24.00}     & \multicolumn{1}{c}{40.60}  & \multicolumn{1}{c}{38.11} & \multicolumn{1}{c|}{34.06} & \multicolumn{1}{c}{56.45}  & \multicolumn{1}{c}{50.78} & \multicolumn{1}{c|}{45.37} & \multicolumn{1}{c}{49.90}  & \multicolumn{1}{c}{47.11} & \multicolumn{1}{c}{43.73}  \\

\multicolumn{2}{l}{\multirow{1}{*}{DD~\cite{hu2022feature}}}            & \multicolumn{1}{c}{33.31}  & \multicolumn{1}{c}{30.57} & \multicolumn{1}{c|}{28.14}     & \multicolumn{1}{c}{44.66}  & \multicolumn{1}{c}{41.47} & \multicolumn{1}{c|}{37.41} & \multicolumn{1}{c}{57.42}  & \multicolumn{1}{c}{52.69} & \multicolumn{1}{c|}{47.90} & \multicolumn{1}{c}{51.71}  & \multicolumn{1}{c}{48.48} & \multicolumn{1}{c}{44.95}    \\

\multicolumn{2}{l}{\multirow{1}{*}{\textbf{ProtoOT}}}              
& \multicolumn{1}{c}{\textbf{36.75}}& \multicolumn{1}{c}{\textbf{33.58}}     & \multicolumn{1}{c|}{\textbf{31.33}}    & \multicolumn{1}{c}{\textbf{48.93}}    & \multicolumn{1}{c}{\textbf{45.93}}    & \multicolumn{1}{c|}{\textbf{41.59}}    & \multicolumn{1}{c}{\textbf{64.01}}    & \multicolumn{1}{c}{\textbf{59.22}}    & \multicolumn{1}{c|}{\textbf{54.50}}    & \multicolumn{1}{c}{\textbf{54.85}}    & \multicolumn{1}{c}{\textbf{53.49}}    &  \multicolumn{1}{c}{\textbf{51.37}}     \\

\midrule

\multicolumn{2}{c}{\multirow{2}{*}}  & \multicolumn{3}{c|}{Product{$\rightarrow$}Clipart} & \multicolumn{3}{c|}{Clipart{$\rightarrow$}Product}  & \multicolumn{3}{c|}{Art{$\rightarrow$}Clipart} & \multicolumn{3}{c}{Clipart{$\rightarrow$}Art}   \\
\multicolumn{2}{c}{}                                          & \multicolumn{1}{c}{P@1} & \multicolumn{1}{c}{P@5}  & \multicolumn{1}{c|}{P@15} & \multicolumn{1}{c}{P@1} & \multicolumn{1}{c}{P@5}  & \multicolumn{1}{c|}{P@15} & \multicolumn{1}{c}{P@1} & \multicolumn{1}{c}{P@5}  & \multicolumn{1}{c|}{P@15} & \multicolumn{1}{c}{P@1} & \multicolumn{1}{c}{P@5}  & \multicolumn{1}{c}{P@15}
\\ 
\midrule

\multicolumn{2}{l}{ID~\cite{wu2018unsupervised}}                         & \multicolumn{1}{c}{31.52}  & \multicolumn{1}{c}{28.55} & \multicolumn{1}{c|}{24.15}     & \multicolumn{1}{c}{24.01}  & \multicolumn{1}{c}{22.42} & \multicolumn{1}{c|}{20.60} & \multicolumn{1}{c}{26.78}  & \multicolumn{1}{c}{24.79} & \multicolumn{1}{c|}{21.64} & \multicolumn{1}{c}{21.17}  & \multicolumn{1}{c}{17.86} & \multicolumn{1}{c}{14.71}   \\ 

\multicolumn{2}{l}{\multirow{1}{*}{ProtoNCE~\cite{li2020prototypical}}}               & \multicolumn{1}{c}{36.13}  & \multicolumn{1}{c}{33.99} & \multicolumn{1}{c|}{28.24}     & \multicolumn{1}{c}{21.17}  & \multicolumn{1}{c}{20.63} & \multicolumn{1}{c|}{20.47} & \multicolumn{1}{c}{28.97}  & \multicolumn{1}{c}{26.15} & \multicolumn{1}{c|}{22.98} & \multicolumn{1}{c}{21.33}  & \multicolumn{1}{c}{17.40} & \multicolumn{1}{c}{14.46}  \\

\multicolumn{2}{l}{\multirow{1}{*}{CDS~\cite{kim2021cds}}}                & \multicolumn{1}{c}{37.69}  & \multicolumn{1}{c}{34.99} & \multicolumn{1}{c|}{30.42}     & \multicolumn{1}{c}{27.24}  & \multicolumn{1}{c}{26.46} & \multicolumn{1}{c|}{24.86} & \multicolumn{1}{c}{25.59}  & \multicolumn{1}{c}{23.77} & \multicolumn{1}{c|}{22.41} & \multicolumn{1}{c}{22.41}  & \multicolumn{1}{c}{20.34} & \multicolumn{1}{c}{17.34}  \\

\multicolumn{2}{l}{\multirow{1}{*}{PCS~\cite{yue2021prototypical}}}               & \multicolumn{1}{c}{39.51}  & \multicolumn{1}{c}{37.51} & \multicolumn{1}{c|}{32.81}     & \multicolumn{1}{c}{26.39}  & \multicolumn{1}{c}{25.86} & \multicolumn{1}{c|}{24.92} & \multicolumn{1}{c}{31.23}  & \multicolumn{1}{c}{28.74} & \multicolumn{1}{c|}{26.11} & \multicolumn{1}{c}{24.51}  & \multicolumn{1}{c}{21.27} & \multicolumn{1}{c}{17.54}  \\

\multicolumn{2}{l}{\multirow{1}{*}{DD~\cite{hu2022feature}}}              & \multicolumn{1}{c}{42.26}  & \multicolumn{1}{c}{37.42} & \multicolumn{1}{c|}{33.74}     & \multicolumn{1}{c}{27.79}  & \multicolumn{1}{c}{27.26} & \multicolumn{1}{c|}{25.97} & \multicolumn{1}{c}{32.67}  & \multicolumn{1}{c}{30.79} & \multicolumn{1}{c|}{28.70} & \multicolumn{1}{c}{27.26}  & \multicolumn{1}{c}{23.94} & \multicolumn{1}{c}{20.53}    \\

\multicolumn{2}{l}{\multirow{1}{*}{\textbf{ProtoOT}}}              
 & \multicolumn{1}{c}{\textbf{44.76}}& \multicolumn{1}{c}{\textbf{42.64}}     & \multicolumn{1}{c|}{\textbf{38.83}}    & \multicolumn{1}{c}{\textbf{29.92}}    & \multicolumn{1}{c}{\textbf{30.15}}    & \multicolumn{1}{c|}{\textbf{29.29}}    & \multicolumn{1}{c}{\textbf{35.39}}    & \multicolumn{1}{c}{\textbf{34.57}}    & \multicolumn{1}{c|}{\textbf{32.05}}    & \multicolumn{1}{c}{\textbf{28.96}}    & \multicolumn{1}{c}{\textbf{25.58}}    &  \multicolumn{1}{c}{\textbf{22.21}}     \\ 

\midrule
\bottomrule
\end{tabular}
}
\caption{Unsupervised Cross-domain Retrieval Accuracy (\%) on Office-Home}
\label{tab:acc on office-home}
\end{table*}
\subsection{A Unified ProtoOT Framework for UCIR}
In UCIR, both the intra-domain representation learning and cross-domain alignment are based on ProtoOT. Furthermore, the two are dedicated to enhancing the local consistency and global discriminability of features through the same form of contrastive loss. Therefore, the two parts of the UCIR task can be unified within the framework based on ProtoOT. In summary, our total training objective can be written as:

\begin{equation}{\label{loss_total}}
    \mathcal{L}_{total} = \mathcal{L}_{in} + \lambda \mathcal{L}_{cr}
\end{equation}

\section{Experiments}
\subsection{Datasets}
We evaluate our proposed method on two datasets: Office-Home and DomainNet. The Office-Home \cite{venkateswara2017deep} dataset comprises 4 domains (Art, Clipart, Product, Real) encompassing 65 categories. For our experiments, we employ all available images. The DomainNet \cite{peng2019moment} dataset consists of 6 domains (Clipart, Infograph, Painting, Quickdraw, Real, and Sketch). Based on the criteria established by \cite{hu2022feature}, we utilize 7 categories with over 200 images in each domain for our experiments.

\begin{table*}[htpb]
\centering
{
\begin{tabular}{cccccccccccccc}
\toprule
\midrule

\multicolumn{1}{l}{}&\multicolumn{1}{l}{Average} &\multicolumn{2}{c}{ID} & \multicolumn{2}{c}{ProtoNCE}  & \multicolumn{2}{c}{CDS} & \multicolumn{2}{c}{PCS}  &\multicolumn{2}{c}{DD} & \multicolumn{1}{c|}{\textbf{ProtoOT}}  & \multicolumn{1}{c}{Improvement}\\ 
\midrule

\multicolumn{1}{l}{}&\multicolumn{1}{l}{P@50}   & \multicolumn{2}{c}{37.07} & \multicolumn{2}{c}{37.21}  & \multicolumn{2}{c}{36.89}  & \multicolumn{2}{c}{41.23} & \multicolumn{2}{c}{47.09}     & \multicolumn{1}{c|}{\textbf{64.37}}  & \multicolumn{1}{c}{\textbf{+17.28}} 
\\ 

\multicolumn{1}{l}{\multirow{1}{*}{DomainNet}}&\multicolumn{1}{l}{P@100}  & \multicolumn{2}{c}{33.34} & \multicolumn{2}{c}{32.59} & \multicolumn{2}{c}{31.84}  & \multicolumn{2}{c}{36.87} & \multicolumn{2}{c}{43.47} & \multicolumn{1}{c|}{\textbf{61.93}}   & \multicolumn{1}{c}{\textbf{+18.46}} \\ 

\multicolumn{1}{l}{}&\multicolumn{1}{l}{P@200}  & \multicolumn{2}{c}{28.72} & \multicolumn{2}{c}{27.85}  & \multicolumn{2}{c}{26.69}  & \multicolumn{2}{c}{31.74} & \multicolumn{2}{c}{39.09} & \multicolumn{1}{c|}{\textbf{57.26}}   & \multicolumn{1}{c}{\textbf{+18.17}} \\
\midrule

\multicolumn{1}{l}{}&\multicolumn{1}{l}{P@1}   & \multicolumn{2}{c}{33.18} & \multicolumn{2}{c}{35.49}  & \multicolumn{2}{c}{37.17}  & \multicolumn{2}{c}{38.07} & \multicolumn{2}{c}{40.67}     & \multicolumn{1}{c|}{\textbf{43.85}}  & \multicolumn{1}{c}{\textbf{+3.18}} 
\\ 

\multicolumn{1}{l}{\multirow{1}{*}{Office-Home}}&\multicolumn{1}{l}{P@5}  & \multicolumn{2}{c}{29.76} & \multicolumn{2}{c}{32.15} & \multicolumn{2}{c}{34.63}  & \multicolumn{2}{c}{35.01} & \multicolumn{2}{c}{37.60} & \multicolumn{1}{c|}{\textbf{41.29}}    & \multicolumn{1}{c}{\textbf{+3.69}} \\

\multicolumn{1}{l}{}&\multicolumn{1}{l}{P@15}  & \multicolumn{2}{c}{25.78} & \multicolumn{2}{c}{28.30}  & \multicolumn{2}{c}{31.30}  & \multicolumn{2}{c}{31.42} & \multicolumn{2}{c}{34.15} & \multicolumn{1}{c|}{\textbf{37.98}}   & \multicolumn{1}{c}{\textbf{+3.83}}\\

\midrule
\bottomrule
\end{tabular}
}
\caption{Average Accuracy (\%) for Unsupervised Cross-domain Retrieval on DomainNet and Office-Home }
\label{tab: avg acc}
\end{table*}
\subsection{Implementation Details}
We employ the ResNet-50\cite{he2016deep} architecture as the encoder $f_{\theta}$, initializing its parameters using MoCov2\cite{chen2020improved} trained on the unlabeled ImageNet dataset\cite{deng2009imagenet}. The extracted features undergo $l_2$-normalization. Our optimization employs the Adam optimizer with a learning rate of $2.5 \times 10^{-4}$ over 200 epochs, with a batch size of 64. The K-means clustering is performed on the memory bank at the end of each epoch. For the Sinkhorn Algorithm\cite{cuturi2013sinkhorn}, the entropic regularization coefficient $\epsilon$ is set to 0.05 and following \cite{caron2020unsupervised} the iterations is 3. The number of prototypes corresponds to the number of classes in the training set: 65 for Office-Home and 7 for DomainNet. In the initial phase of training, we employ the same loss function as used in MoCov2\cite{chen2020improved} for per-domain pre-training.

\subsection{Evaluation Metrics}
Follow \cite{hu2022feature}, our evaluation metric involves measuring precision across the top 1/5/15 retrieved images for the Office-Home dataset, considering a minimum of 15 images per category. For the subset of the selected 7 categories within DomainNet, we assess precision using the top 50/100/200 retrieved images. Our primary focus lies in category-level cross-domain retrieval. In this context, we consider retrieved images that correspond to the same semantic classes as the query as correct matches.

\subsection{Baselines}

In our experimental validation, we assess the effectiveness of ProtoOT by comparing it against the following baseline methods:
\textbf{ID} \cite{wu2018unsupervised}: This approach learns instance-level information by discerning distinctions between instances.
\textbf{ProtoNCE} \cite{li2020prototypical}: This method introduces the integration of prototypes to effectively capture semantic information and enhance cluster capabilities.
\textbf{CDS} \cite{kim2021cds}: Taking both intra- and cross-domain discriminative aspects into consideration among instances, this method aims to address domain adaptation.
\textbf{PCS} \cite{yue2021prototypical}: Designed for few-shot unsupervised domain adaptation, this cross-domain self-supervised learning method employs K-means to assign prototypes and utilizes similarity vectors for cross alignment.
\textbf{DD} \cite{hu2022feature}: This method proposes a distance-of-distance loss for quantifying domain discrepancy.
These baselines serve as benchmarks for validating the performance of our ProtoOT.

\subsection{Comparison with State-of-the-Art Methods}

ProtoOT's impressive performance is evident from the summarized results in Table \ref{tab: avg acc}. On the Office-Home dataset, ProtoOT outperforms the leading baseline\cite{hu2022feature} with average improvements of 3.18\%, 3.69\%, and 3.83\% in P@1, P@5, and P@15, respectively. The improvements are even more pronounced on DomainNet, showcasing substantial enhancements of 17.28\%, 18.46\%, and 18.17\% in P@50, P@100, and P@200.
Table \ref{tab:acc on domainnet} and \ref{tab:acc on office-home} show the detailed retrieval performance for Office-Home and DomainNet, respectively.  It is worth noting that the Painting $\rightarrow$ Clipart retrieval on DomainNet shows the most significant improvement, with an accuracy boost of 30.54\% in terms of P@100. Furthermore, ProtoOT maintains over 50\% retrieval precision consistently, even in challenging tasks that involve Quickdraw, with an impressive average enhancement of 20.40\%, 20.51\%, and 20.47\% in P@50, P@100, and P@200, respectively. This is attributed to the strong matching capabilities of ProtoOT.

\subsection{Ablation Study}

To thoroughly assess the effectiveness of our proposed ProtoOT, we meticulously compare its impact on both intra-domain feature representation learning and cross-domain alignment, as outlined in Table \ref{tab: ablation1}. We systematically replace each component with the standard OT to demonstrate the superiority of our model's design.
The experimental results clearly demonstrate the substantial contribution of cross-domain alignment, highlighting the significance of bridging the domain gap between distinct domains for the UCIR task. Furthermore, it's evident that ProtoOT maintains its advantageous characteristics.
In summary, the unified ProtoOT framework displays remarkable performance in both intra-domain clustering and cross-domain alignment, underscoring its exceptional effectiveness.

\begin{table}[htpb]
\centering
\begin{tabular}{ccccc}
\toprule
 \multicolumn{1}{c}{Intra}& \multicolumn{1}{c}{Cross}& \multicolumn{1}{c}{P@50}  & \multicolumn{1}{c}{P@100}          &\multicolumn{1}{c}{P@200}\\ 
\midrule

\multicolumn{1}{c}{\text{SOT}}   &\XSolidBrush &  \multicolumn{1}{c}{55.72}    &    \multicolumn{1}{c}{52.41} & \multicolumn{1}{c}{46.58}
\\ 
\multicolumn{1}{c}{\text{SOT}} &\multicolumn{1}{c}{\text{SOT}}&    
 \multicolumn{1}{c}{60.08}    &    \multicolumn{1}{c}{57.01} & \multicolumn{1}{c}{51.46}
\\ 
\multicolumn{1}{c}{\text{SOT}}  & \multicolumn{1}{c}{\text{ProtoOT}} & \multicolumn{1}{c}{\text{60.58}}    &    \multicolumn{1}{c}{\text{57.78}} & \multicolumn{1}{c}{\text{52.16}}     
\\
\multicolumn{1}{c}{\text{ProtoOT}}   &\XSolidBrush & \multicolumn{1}{c}{56.51}  &  \multicolumn{1}{c}{\text{53.13}}  &   \multicolumn{1}{c}{\text{47.02}} 
\\ 
\multicolumn{1}{c}{\text{ProtoOT}} &\multicolumn{1}{c}{\text{SOT}}&  
\multicolumn{1}{c}{62.83}    &    \multicolumn{1}{c}{60.29} & \multicolumn{1}{c}{55.43}                 
\\ 
\multicolumn{1}{c}{\text{ProtoOT}}  & \multicolumn{1}{c}{\text{ProtoOT}} & \multicolumn{1}{c}{\textbf{64.37}}    &    \multicolumn{1}{c}{\textbf{61.93}} & \multicolumn{1}{c}{\textbf{57.26}}     
\\
\bottomrule                   
\end{tabular}

\caption{
Evaluation of the effectiveness of ProtoOT on both intra-domain feature representation learning and cross-domain alignment (Average on 12 tasks). SOT is an abbreviation for the standard Optimal Transport.
} 
\label{tab: ablation1}

\end{table}

\begin{figure*}[htpb]
\centering
\includegraphics[width=0.7\textwidth]{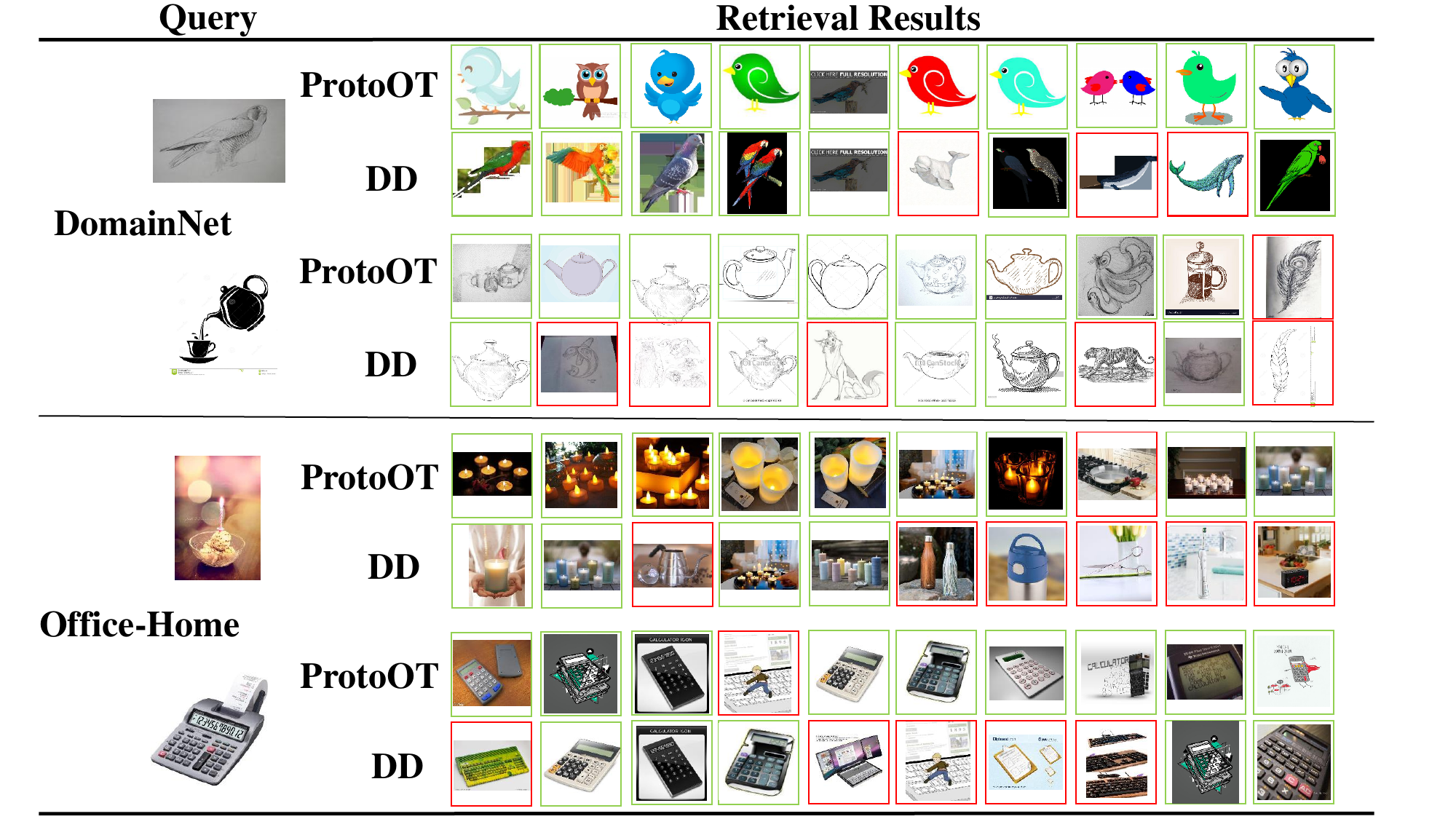}
\caption{
{Top 10 retrieval results in DomainNet and Office-Home. The green and red boxes denote correct and incorrect retrievals, respectively.}
}
\label{fig: Top 10 retrieval results.}
\end{figure*}

To further validate the superiority of our ProtoOT, we conducted a comparison with several common variants of OT. The Unbalanced OT\cite{sejourne2022unbalanced} uses soft penalties to relax the conservation of marginal constraints.
The Partial OT\cite{figalli2010optimal,phatak2022computing} focuses on transporting only a fraction of mass at the lowest cost. Due to their intrinsic characteristics that are not well suited for the UCIR task, as depicted in Table \ref{tab: ablation_ot}, they yield suboptimal performance.

\begin{table}[htbp]
\centering
\begin{tabular}{ccccc}
\toprule
 \multicolumn{1}{c}{Intra}& \multicolumn{1}{c}{Cross}& \multicolumn{1}{c}{P@50}  & \multicolumn{1}{c}{P@100}          &\multicolumn{1}{c}{P@200}\\ 
\midrule

\multicolumn{1}{c}{\text{UOT}} &\multicolumn{1}{c}{\text{UOT}}&  
\multicolumn{1}{c}{60.23}    &    \multicolumn{1}{c}{57.39} & \multicolumn{1}{c}{52.47}                 
\\ 
\multicolumn{1}{c}{\text{POT}} &\multicolumn{1}{c}{\text{POT}}&  
\multicolumn{1}{c}{60.61}    &    \multicolumn{1}{c}{57.93} & \multicolumn{1}{c}{53.23}                 
\\ 
\multicolumn{1}{c}{\text{ProtoOT}}  & \multicolumn{1}{c}{\text{ProtoOT}} & \multicolumn{1}{c}{\textbf{64.37}}    &    \multicolumn{1}{c}{\textbf{61.93}} & \multicolumn{1}{c}{\textbf{58.26}}     
\\
\bottomrule                   
\end{tabular}

\caption{
The retrieval performance comparison for several OT variants(Average on 12 tasks on DomainNet). UOT and POT represent Unbalanced OT and Partial OT, respectively.
} 
\label{tab: ablation_ot}

\end{table}

To investigate the impact of the coefficient $\lambda$ in Eq.(\ref{loss_total}), we conduct  hyper-parameter analysis experiments on DomainNet as shown in Table \ref{tab: lambda}. The favorable performance can be obtained within the range of 0.001 to 0.1. Based on the observation, we set $\lambda = 0.01$ in all experiments.

\begin{table}[htpb]
\centering
\begin{tabular}{ccccccc}
\toprule
 \multicolumn{1}{c}{$\lambda$} &\multicolumn{1}{c}{P@50}  & \multicolumn{1}{c}{P@100}          &\multicolumn{1}{c}{P@200}\\ 
\midrule
\multicolumn{1}{c}{0.001}  &\multicolumn{1}{c}{61.56}&  \multicolumn{1}{c}{59.30}    &    \multicolumn{1}{c}{54.49} \\

\multicolumn{1}{c}{0.005} &\multicolumn{1}{c}{\text{62.40}}&  \multicolumn{1}{c}{\text{59.83}}    &    \multicolumn{1}{c}{\text{54.74}} \\
\multicolumn{1}{c}{0.01} &\multicolumn{1}{c}{\textbf{64.37}}&  \multicolumn{1}{c}{\textbf{61.93}}    &    \multicolumn{1}{c}{\textbf{57.26}} \\
\multicolumn{1}{c}{0.05} &\multicolumn{1}{c}{\text{62.93}}&  \multicolumn{1}{c}{\text{59.91}}    &    \multicolumn{1}{c}{\text{54.81}} \\
\multicolumn{1}{c}{0.1} &\multicolumn{1}{c}{\text{62.48}}&  \multicolumn{1}{c}{\text{59.41}}    &    \multicolumn{1}{c}{\text{54.28}} \\

\bottomrule                   
\end{tabular}
\caption{
The effect of coefficent $\lambda$ in equation 12 (Average on 12 tasks on DomainNet)
} 
\label{tab: lambda}
\end{table}

\subsection{Visualization of the Retrieval Results}

We offer a quantitative visualization of our proposed ProtoOT alongside the most promising state-of-the-art method, DD\cite{hu2022feature}, by showcasing cross-domain retrieval outcomes. The results presented in Figure \ref{fig: Top 10 retrieval results.}, ordered from top to bottom, correspond to the retrieval results of the following four tasks: Sketch-Clipart, Clipart-Sketch, Art-Product, and Product-Art cross-domain retrieval.  Upon observing the outcomes, it becomes evident that out ProtoOT is effective in eliminating the interference caused by samples that share visual similarity but belong to different categories.

\section{Conclusion} 

In this paper, we have addressed the challenge of unsupervised cross-domain image retrieval (UCIR) by introducing ProtoOT, a novel Optimal Transport formulation explicitly designed for this task.  We have shown that by unifying the intra-domain representation learning and the cross-domain feature alignment within the ProtoOT framework, we can capitalize on their synergistic potential and significantly enhance UCIR performance.  Our experimental results have demonstrated the superiority of ProtoOT over existing state-of-the-art techniques across benchmark datasets. 

\section{Acknowledgments}
This work was supported by Shanghai Local College Capacity Building Program (23010503100), Shanghai Sailing Program (21YF1429400, 22YF1428800), NSFC (No.62303319),  Shanghai Frontiers Science Center of Human-centered Artificial Intelligence (ShangHAI), MoE Key Laboratory of Intelligent Perception and Human-Machine Collaboration (ShanghaiTech University), Shanghai Clinical Research and Trial Center and
Shanghai Engineering Research Center of Intelligent Vision and Imaging.


\newpage

\section{Supplementary materials}

\subsection{Pre-training Details}
In the initial phase of training, we employ the same loss function as used in MoCov2\cite{chen2020improved} for per-domain pre-training. The loss of the pre-training stage is as follow:
\begin{equation}
    \mathcal{L}_{pre} =\sum\limits_{i \in \mathcal{D}}
    -\log\frac{\exp(\bm{q}^T_i\bm{q}'_i/\tau)}{\sum\limits_{\bm{j}\in \mathcal{D}} \exp(\bm{q}^T_i\bm{q}'_j/\tau)},
\end{equation}
\noindent where $\bm{q}_i $ and $\bm{q}'_i $  can be the image features from different augmented views of image $i$, $\tau$ is the temperature value. 
Specifically, during the initial 30 epochs of training, the pre-training loss term is incorporated to facilitate feature learning. Beyond the 30th epoch, this instance-wise loss is omitted to promote the acquisition of the learned features with more distinct class-level semantic separability.

\subsection{Visualization}
\begin{figure*}[!h]
     \centering
     \begin{subfigure}[b]{0.47\textwidth}
         \centering
         \includegraphics[width=\textwidth]{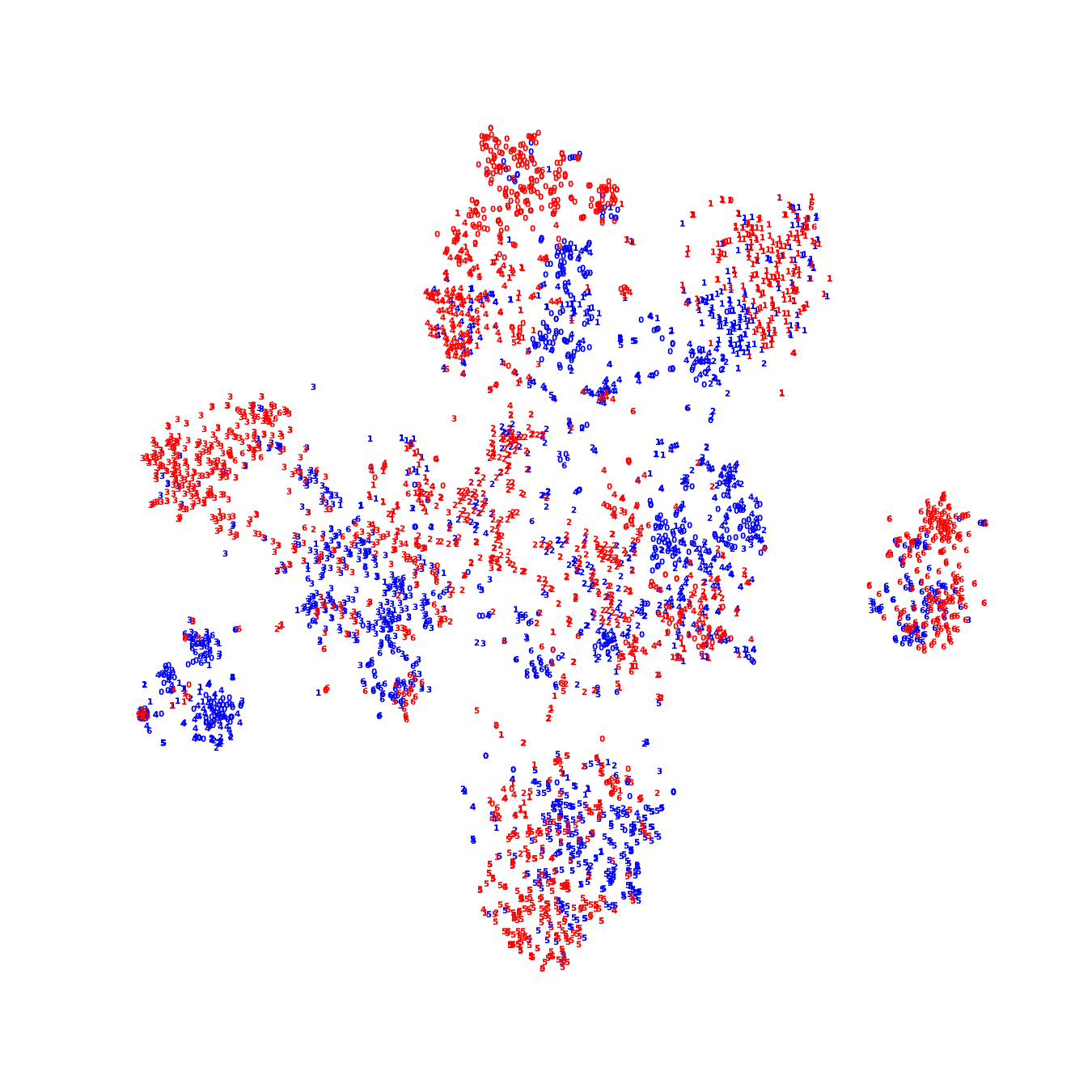}
         \caption{DD (Clipart-Sketch)}
     \end{subfigure}
     \hfill
     \begin{subfigure}[b]{0.47\textwidth}
         \centering
         \includegraphics[width=\textwidth]{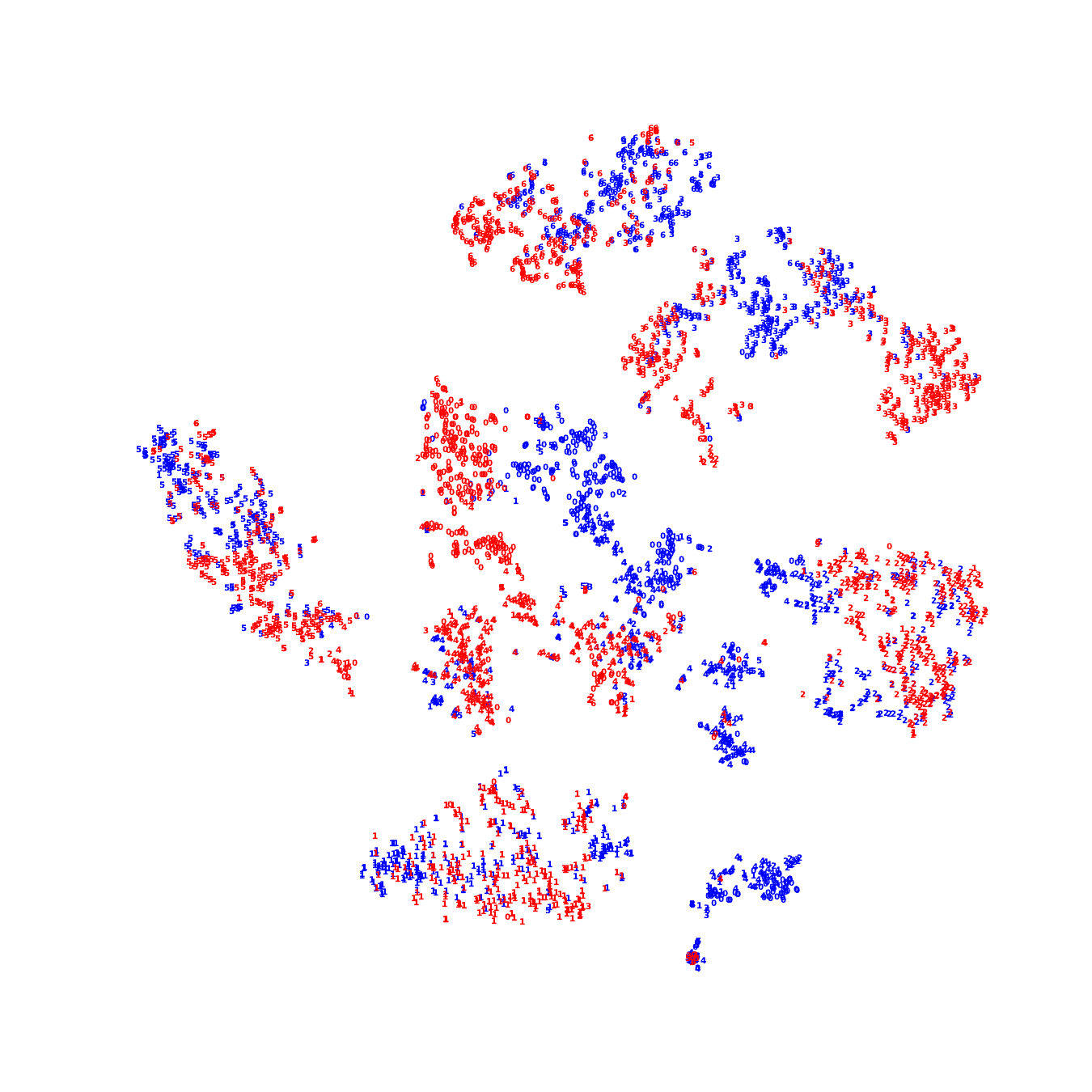}
         \caption{ProtoOT (Clipart-Sketch)}
     \end{subfigure}\\
     \begin{subfigure}[b]{0.47\textwidth}
         \centering
         \includegraphics[width=\textwidth]{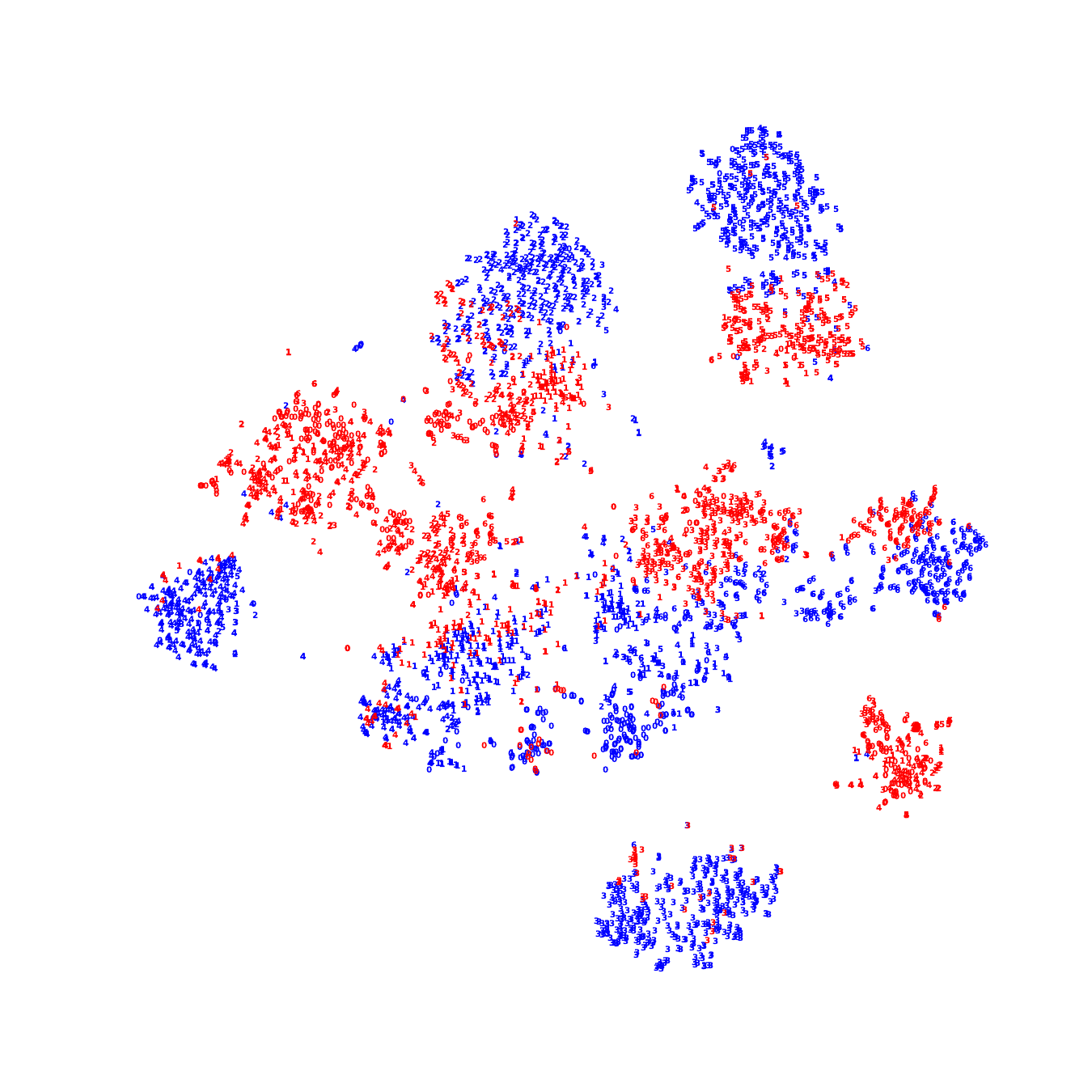}
         \caption{DD (Painting-Clipart)}
     \end{subfigure}
     \hfill
     \begin{subfigure}[b]{0.47\textwidth}
         \centering
         \includegraphics[width=\textwidth]{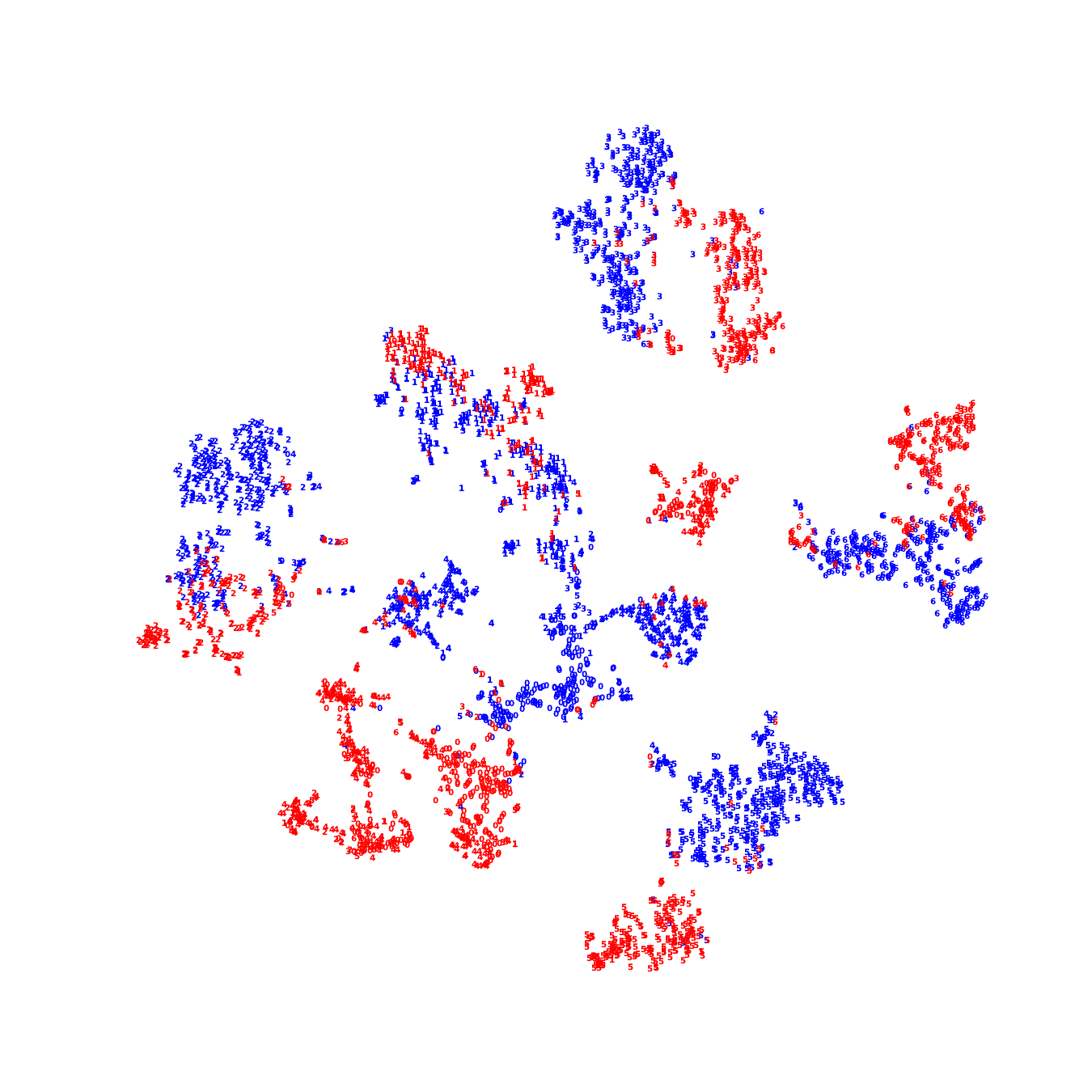}
         \caption{ProtoOT (Painting-Clipart)}
     \end{subfigure}
    \caption{2-d t-SNE visualizations feature representations learned by DD and our proposed ProtoOT on DomainNet. Each class is represented by a number and Each sample is colored by its corresponding domain.}
    \label{fig:tsne}
\end{figure*}

\begin{figure*}[htpb]
\centering
\includegraphics[width=1\textwidth]{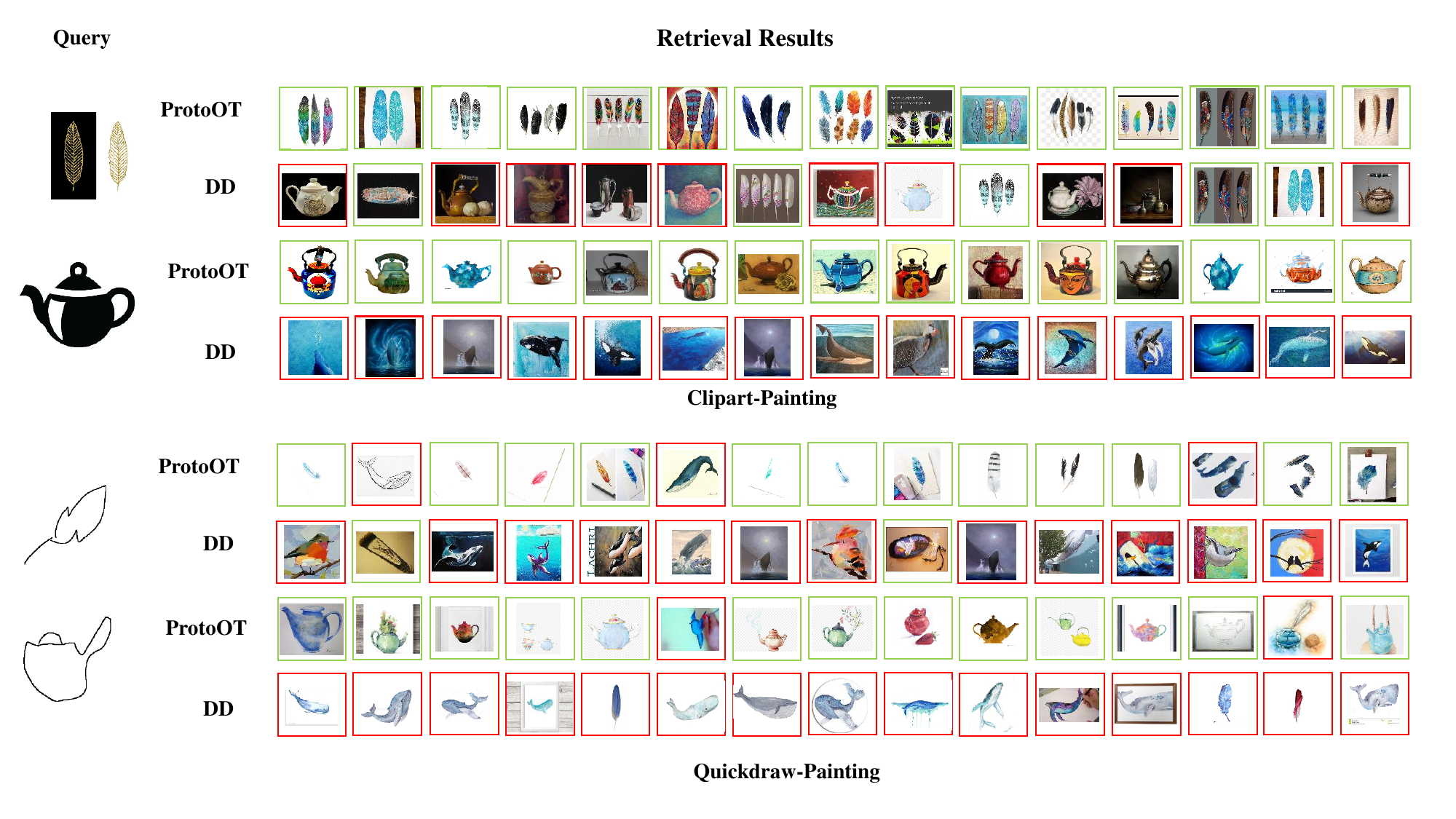}
\caption{
{Top 15 retrieval results in DomainNet. The green and red boxes denote correct and incorrect retrievals, respectively.}
}
\label{fig: Top 15 domainnet.}
\end{figure*}

\begin{figure*}[htpb]
\centering
\includegraphics[width=1\textwidth]{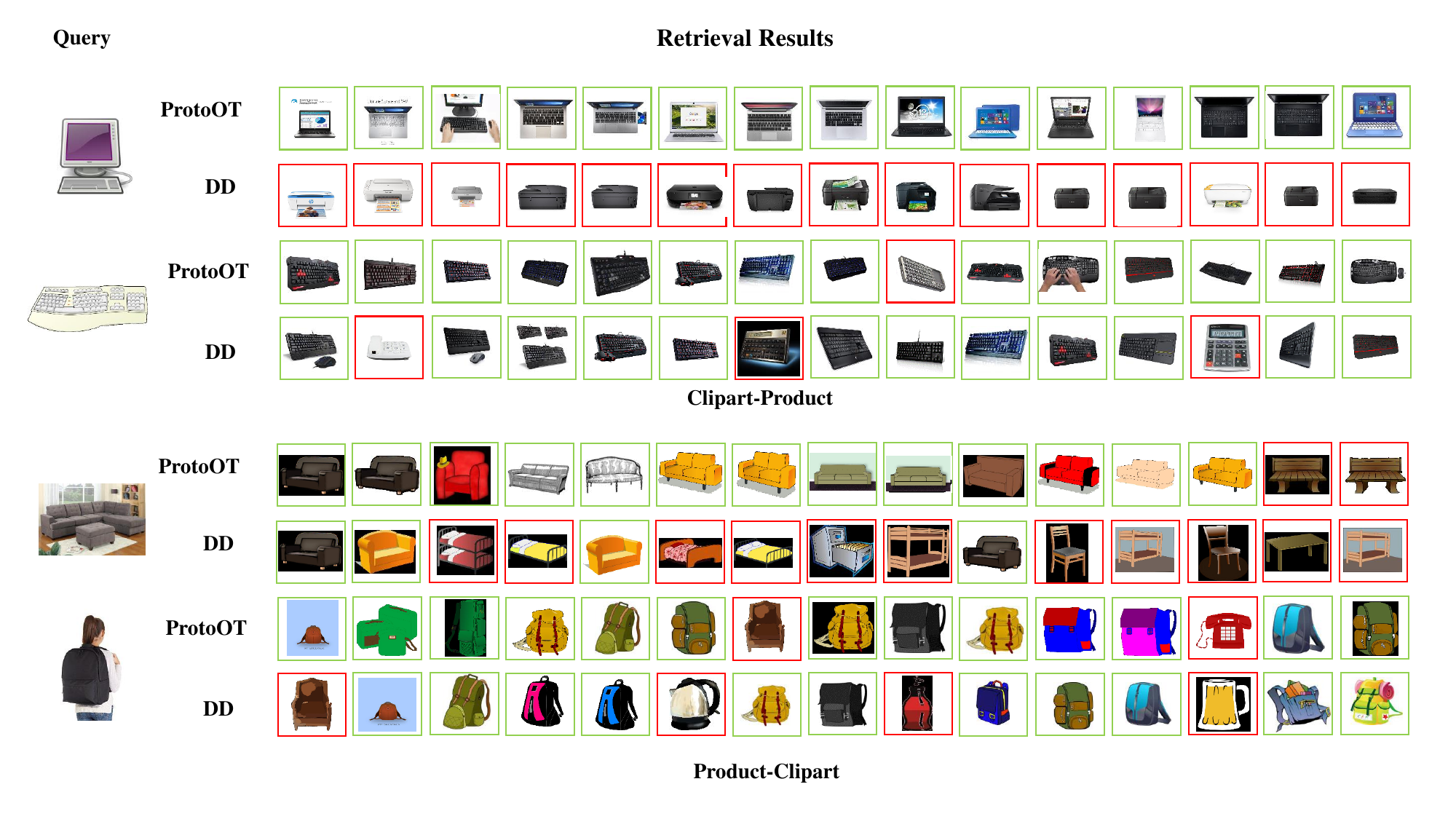}
\caption{
{Top 15 retrieval results in Office-Home. The green and red boxes denote correct and incorrect retrievals, respectively.}
}
\label{fig: Top 15 office-home.}
\end{figure*}

To provide a more intuitive comparison between our ProtoOT and DD\cite{hu2022feature}, we visualize Clipart-Sketch and Painting-Clipart on DomainNet. As shown in Figgure \ref{fig:tsne}, we can observe that compared to DD, our ProtoOT yields more compact intra-class features and distinct inter-class separations.This can eliminate interference from other categories, thus contributing to the effectiveness of our ProtoOT approach. In Figure \ref{fig: Top 15 domainnet.} and \ref{fig: Top 15 office-home.}, we present additional retrieval results to demonstrate the effectiveness of our ProtoOT.

\end{document}